\documentclass[10pt,twocolumn,letterpaper]{article}
\usepackage[affil-it]{authblk}
\usepackage{wacv}              

\usepackage{graphicx}
\usepackage{amsmath}
\usepackage{amssymb}
\usepackage{booktabs}

\usepackage{calrsfs}
\DeclareMathAlphabet{\pazocal}{OMS}{zplm}{m}{n}

\newcommand{\Lb}{\pazocal{L}}

\usepackage[pagebackref,breaklinks,colorlinks]{hyperref}

\usepackage[capitalize]{cleveref}
\crefname{section}{Sec.}{Secs.}
\Crefname{section}{Section}{Sections}
\Crefname{table}{Table}{Tables}
\crefname{table}{Tab.}{Tabs.}

\def\wacvPaperID{1585} 
\def\confName{WACV}
\def\confYear{2024}

\begin{document}

\title{GC-VTON: Predicting Globally Consistent and Occlusion Aware Local Flows with Neighborhood Integrity Preservation for Virtual Try-on}

\renewcommand*{\Authands}{, }
\renewcommand*{\Affilfont}{\normalsize\normalfont}

\author[1]{Hamza Rawal}
\author[1]{Muhammad Junaid Ahmad}
\author[2]{Farooq Zaman}
\affil[1]{Motive}
\affil[2]{Information Technology University, Lahore, Pakistan}
\affil[ ]{{\tt\small \{hamzarawal, farooqzaman20\}@gmail.com,} \tt\small 
 junaidahmad1998@outlook.com}


\twocolumn[{%
\renewcommand\twocolumn[1][]{#1}%
\maketitle
\begin{center}
    \centering
    \captionsetup{type=figure}
    \includegraphics[width=0.9\textwidth]{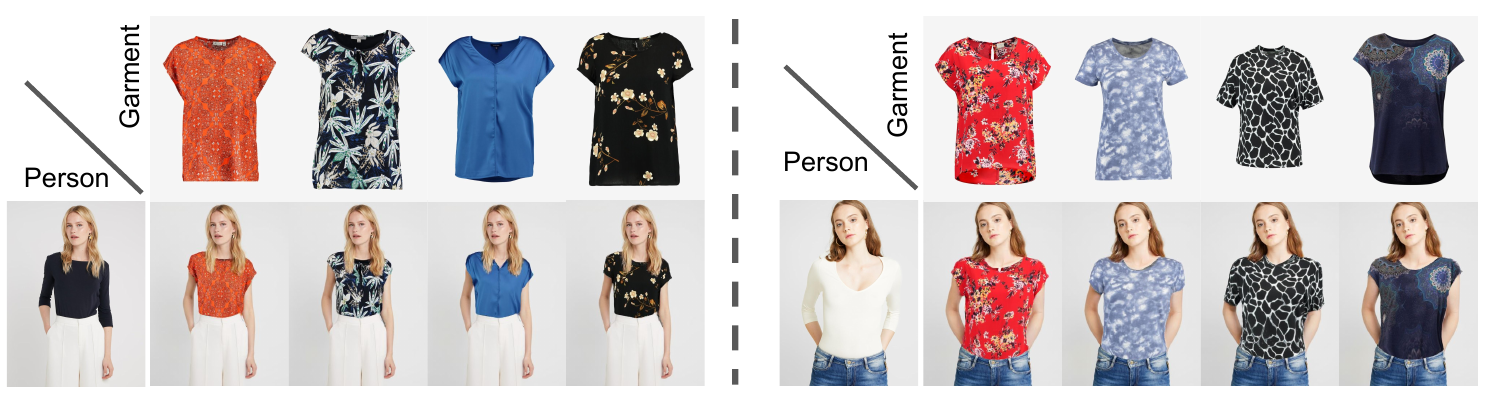}
    \captionof{figure}{Sample try-on images generated by our method over a set of garments with complex textures. Zoom in for details.}
\end{center}%
}]

\maketitle


\begin{abstract}
Flow based garment warping is an integral part of image-based virtual try-on networks. However, optimizing a single flow predicting network for simultaneous global boundary alignment and local texture preservation results in sub-optimal flow fields. Moreover, dense flows are inherently not suited to handle intricate conditions like garment occlusion by body parts or by other garments. Forcing flows to handle the above issues results in various distortions like texture squeezing, and stretching. In this work, we propose a novel approach where we disentangle the global boundary alignment and local texture preserving tasks via our GlobalNet and LocalNet modules. A consistency loss is then employed between the two modules which harmonizes the local flows with the global boundary alignment.
Additionally, we explicitly handle occlusions by predicting body-parts visibility mask, which is used to mask out the occluded regions in the warped garment. The masking prevents the LocalNet from predicting flows that distort texture to compensate for occlusions. We also introduce a novel regularization loss (NIPR), that defines a criteria to identify the regions in the warped garment where texture integrity is violated (squeezed or stretched). NIPR subsequently penalizes the flow in those regions to ensure regular and coherent warps that preserve the texture in local neighborhoods. Evaluation on a widely used virtual try-on dataset demonstrates strong performance of our network compared to the current SOTA methods.

\end{abstract}

\section{Introduction}
\label{sec:intro}
Image-based virtual try-on aims at generating natural, distortion and artifacts-free images of a person wearing a selected garment.
Image synthesis via GANs \cite{goodfellow2014generative} has been widely used in applications like image editing \cite{Wang_2018_CVPR, Shen_2020_CVPR, ling2021editgan}, style-transfer \cite{zhu2017unpaired, Chen_2021_CVPR, Gunawan_2023_CVPR} and image generation \cite{kang2023scaling, zhang2022styleswin, Zhu_2019_CVPR}. However, simply using synthesis methods that holistically change the image does not result in the desired quality in virtual try-on setting. 
Existing methods adopt a scheme where the garment is first warped to meet the target person pose requirements. A GAN based generator network then fuses the warped garment and the target person images to generate a final try-on image.
Traditionally, the warping is either done by a Thin Plate Spline (TPS) warp \cite{han2018viton, yang2020towards, ge2021disentangled, li2021toward, choi2021viton, wang2018toward, issenhuth2020not}, or a dense flow fields based warp \cite{dong2022dressing, he2022style, lee2022high, bai2022single, han2019clothflow}, or a combination of both \cite{yang2021ct}. In any case, the warping is inherently not capable of modeling all the changes that a garment undergoes (e.g occlusions) when it fits on a target person. And forcing it to do so, results in artifacts such as texture squeezing, stretching, and garment tear, etc.

Recently, dense flow networks have shown good warping performance compared to the TPS based warping networks. However, for dense flows, simultaneous alignment of the global boundaries and local texture preservation is still a challenge. While improvement attempts have been made, such as conditioning the flow on global style \cite{he2022style} and sparse/dense body pose \cite{bai2022single, lee2022high}, the results are still far from perfect. We believe that this is a product of over-complicating the job of a flow predicting network and setting unrealistic goals to optimize both the problems simultaneously.
To this end, we propose to disentangle the two jobs by dedicating a separate module to each task. A consistency loss is then utilized between the outputs of the two networks to ensure harmony between local and global flows.

Furthermore, garment-body and garment-garment inter-play introduces various occlusions. For example half sleeves, sleeveless, full sleeves and full neck shirts tend to reveal or conceal certain body parts. The visible body parts can potentially occlude certain garment regions. Forcing flow to model occlusions is an ill-posed problem as the flow can only transform pixel location, but not conceal them. The only way a flow can model occlusions is by tearing the garment and squeezing it around the occlusions.
Additional occlusion may be introduced by garment styles e.g upper garment (shirt) tucked into the bottom garment (pants). Flow predicting networks match global boundaries of this style by predicting high values for the lower portion of the upper garment, hence introducing extreme squeezing artifacts. 
Although handling each of the above occlusion types is equally crucial to get a good warp, existing works either ignore them altogether \cite{he2022style, han2018viton, issenhuth2020not}, or settle for just handling some of them \cite{choi2021viton, han2019clothflow, yang2020towards}. This sets unrealistic goals for the flow predicting network and subsequently results in sub-par try-on results.
To address these issues, we propose to predict a body-parts visibility mask that explicitly masks out the occluded regions from the warped garment.
This setup prevents the flow predicting network from predicting high flow values in the occluded regions, which in turn prevents distortion and artifacts.
We input the visibility masks to the generator module as well, which serves as an added guidance for synthesizing visible skin regions.

\begin{figure*}
\centering
  \includegraphics[width=0.83\textwidth]{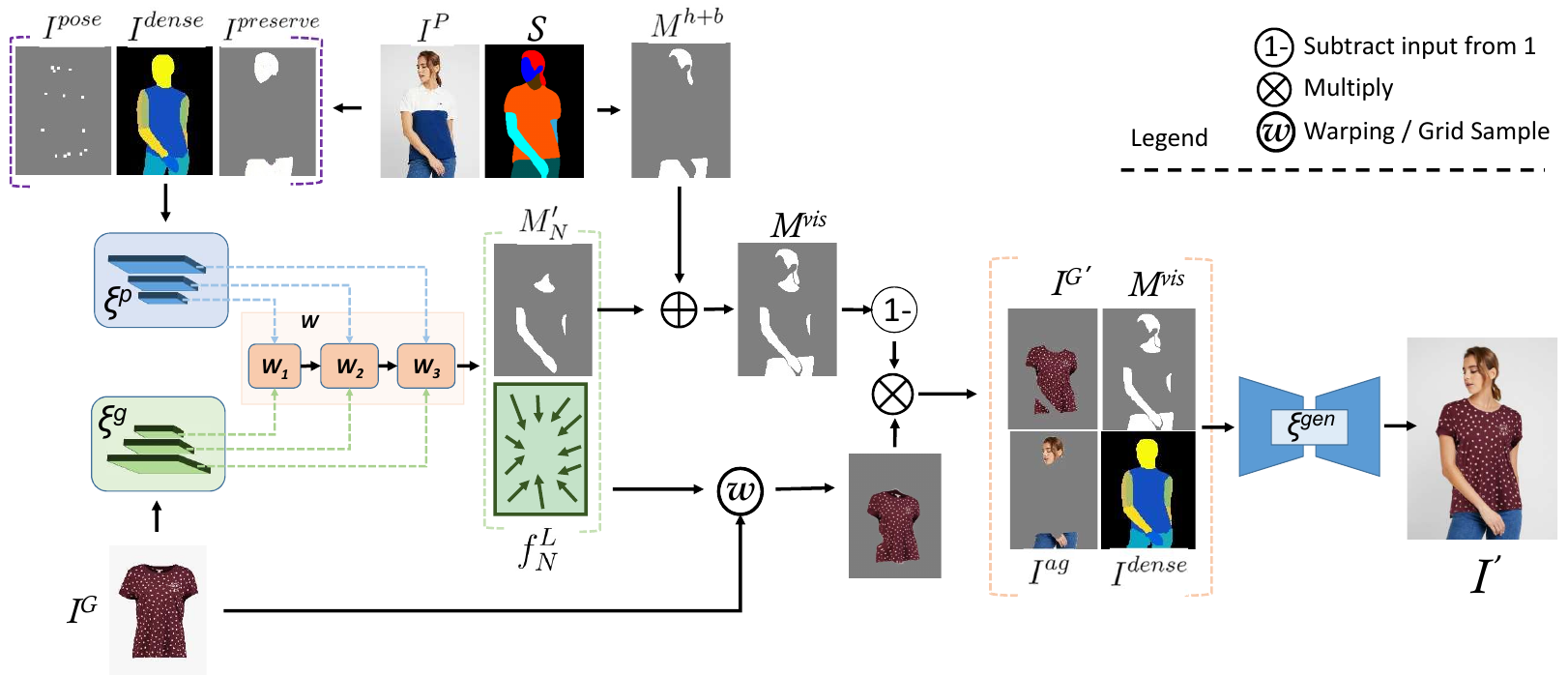}
\caption{An overview of GC-VTON architecture. Person representation maps and garment image are encoded and then fed to the warping module $W=\{W_{1}, W_{2}, W_{3}\}$ which predicts flow fields and body-parts visibility masks. Each warping module $W_{i}$ is composed of a LocalNet and GlobalNet modules. Visibility masks handle occlusions in the warped garment to give $I^{G'}$. $I^{G'}$ and other person representations are input to a generator module to get the try-on image $I^{'}$.} 
\label{fig:model}
\end{figure*}

Additionally, in virtual try-on methods, different losses such as TV-loss \cite{liu2019image} and second-order smoothness loss \cite{ge2021parser} are utilized to prevent irregular flows by enforcing smooth distances between pixels in local neighborhoods.
However, changes in local neighborhoods are governed by global changes in the garment, which these losses fail to take into consideration.
Additionally, they cannot identify the types of the artifacts (squeezing and stretching), thereby fail to apply appropriate penalties on the network. We show that these serious limitations allow artifacts to slip away and result in unrealistic try-on synthesis.
To address these limitations, we propose a novel Neighborhood Integrity Preserving Regularization loss (NIPR). The penalty term in NIPR is globally-informed and artifact-specific, thus penalizing bad warps appropriately.
Our contributions can be summarized as:
\begin{itemize}
    \item We disentangle the global boundaries alignment and local flow adjustment tasks to achieve globally consistent local warps. Utilizing a consistency loss between the outputs of the two modules enforces the necessary harmony.
    \item We propose to estimate body-parts visibility mask to take care of occlusions in the warped garment.
    \item We show the limitations of the existing flow smoothening losses and propose NIPR to effectively guard against texture integrity violation.
\end{itemize}

\section{Related Work}
\subsection{Garment Warping in Virtual Try-on}
Virtual try-on methods using 3D information \cite{he2022style,majithia2022robust,zhao2021m3d,santesteban2021self,patel2020tailornet} have limited applications as the data annotation cost is high. Recently, 2D image based virtual try-on \cite{yang2020towards, han2018viton, he2022style} has gained traction because of its simplicity and less input information required. Generally, a two-stage paradigm is adopted where the first step is to warp the given garment to match the target person pose and then a generator based network fuses the warped garment and the target person image. Initial methods \cite{ge2021disentangled, li2021toward, choi2021viton} applied a TPS based warping technique where a network predicts a set of sparse control points which can be used to warp the garment. Nowadays, dense flow based warping is the method of choice because a flow field has more degree of freedom, thus suited for warping garments with rich textural details \cite{dong2022dressing, lee2022high, han2019clothflow}. A good warp is characterized by two properties: the textural details that it preserves and the global boundary alignment (with the target pose) that it achieves. Although, the two requirements have distinct goals, most of the existing methods expect a single flow predicting network to handle them simultaneously. This leads to unrealistic warps which subsequently results in unnatural try-on results. We propose to handle this by disentangling the two tasks via two separate modules, the outputs of which are harmonized via a consistency loss.

\subsection{Occlusion Handling}
Occlusion is another major complication in virtual try-on which leads to artifacts around the occluded regions. Potential sources of occlusion include body parts and other garments, all of which must be attended to, in order to achieve an artifact-free warp. The existing literature either ignores \cite{he2022style, han2018viton, issenhuth2020not} the occlusion sources altogether or partially attend to some of the sources \cite{choi2021viton, han2019clothflow, yang2020towards} thereby predicting sub-bar warps. In our work, we predict body-parts visibility mask which directly handles the occlusions and prevent flow prediction network from causing distortion.

\subsection{Flow Regularization}
Predicted flows must be smooth (without irregularities) in order to achieve realistic warping results. Multiple losses have been proposed in the existing works to enforce smoothness and coherence in the flow. For instance, TV-Loss \cite{liu2019image, dong2022dressing} minimizes the total variance in a flow field, hence encouraging global spatial smoothness. Authors in \cite{ge2021parser} proposed a second-order smoothness loss, which minimizes a generalized charbonnier loss function \cite{sun2014quantitative} between the distances of a pixel and its two vertical and horizontal neighbors. The intention is to encourage co-linearity in the predicted flow and prevent pixels from falling apart as a result of a bad flow. Authors in \cite{yang2020towards} proposed to equalize the distance of the horizontal and vertical neighbors from a selected point $p$, where $p$ is a member of the set of the predicted control points used to warp the garment.
Furthermore, they equalize the slopes of the lines between the neighbors. The loss encourages the warp to maintain co-linearity, parallelism, and immutability properties of the affine warp.

While these losses have certainly proved to be useful in the context of virtual try-on, they still come with some limitations. We show that the criteria defined in these losses to penalize a bad flow can still be satisfied even when a true (good) warp is not achieved. This can potentially result in certain artifacts being overlooked. We argue that this behavior can be attributed to two discrepancies.
First, the criteria they define is strictly local in nature, whereas the distance between the pixel neighbors is a function of the global changes (in height and width) of the garment.
Second, they cannot identify the nature of the artifacts i-e they do not know if the artifact is caused by stretching or squeezing.
Both the artifacts are distinct in nature and need appropriate penalties. In our work, we propose a new regularization loss (NIPR) that effectively mitigates these issues  by applying globally-compliant and artifact-specific penalties to the flows.

\section{Methodology}
The overall architecture of our method is given in Fig \ref{fig:model}. In this section we detail the working of each module of our approach, its purpose and the intuition behind it.
\subsection{Problem Setting}
Given a target person image \{$I^{P}\, \epsilon \, \mathbb{R}^{3\times W \times H}$\} wearing an original garment \{$I^{PG}$\}, and a garment image \{$I^{G}\, \epsilon \, \mathbb{R}^{3\times W \times H}$\}, the goals of the virtual try-on are to (i) remove \{$I^{PG}$\} from \{$I^{P}$\}, (ii) modify \{$I^{G}$\} to generate try-on garment \{$I^{G'}$\}, and (iii) generate an output image \{$I'\, \epsilon \, \mathbb{R}^{3\times W \times H}$\}, where the person in \{$I'$\} is wearing the modified garment \{$I^{G'}$\}.
\{$I^{G'}$\} should conform to the person's pose and body settings and simultaneously preserve the texture and design details from \{$I^{G}$\}.
And any traces of the person's original garment \{$I^{PG}$\} should not be present in \{$I'$\}. \{$I'$\} should also preserve the non-garment regions of the person \{$I^{P}$\}. 
Simply put, the generated image \{$I'$\} should look natural and plausible.
Like other methods \cite{he2022style, lee2022high}, we assume that we have access to person pose \{$I^{pose}$\}, densepose \{$I^{dense}$\} and person body segmentation map \{$S$\}.

\subsection{Network Overview}
\begin{figure}
\centering
  \includegraphics[width=0.5\textwidth]{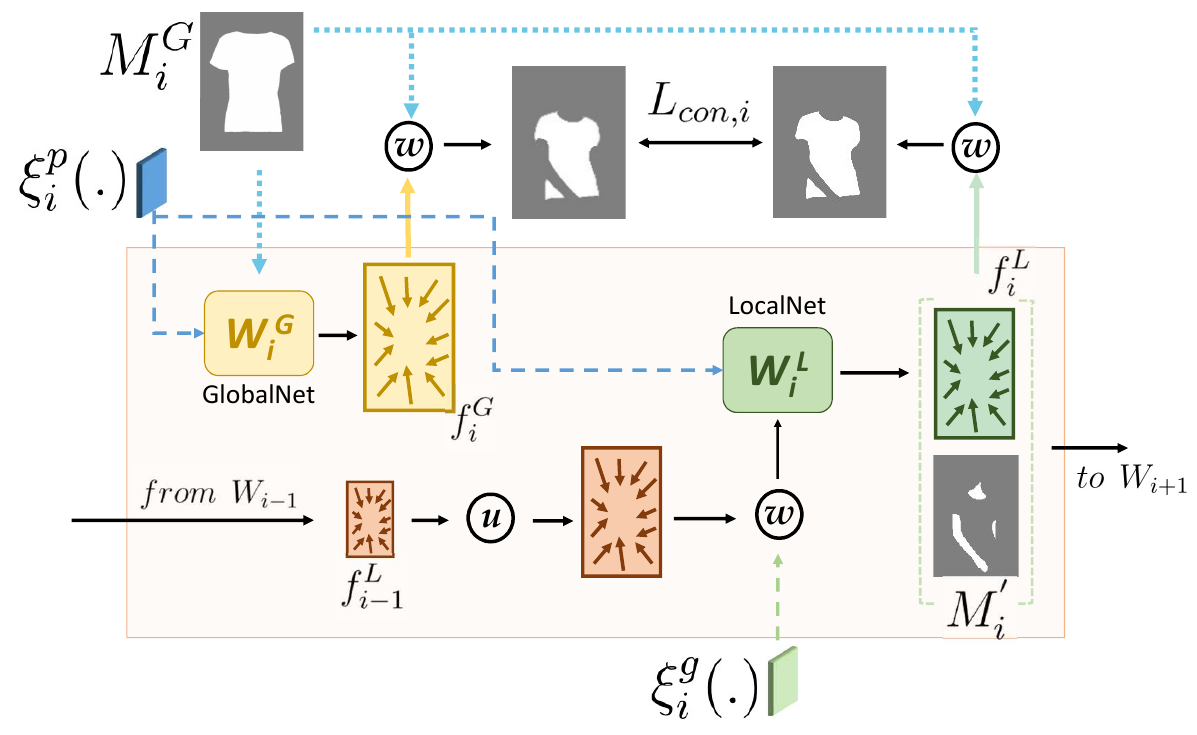}
\caption{Warping block design. Local flow from previous block \{$f_{i-1}^{L}$\} warps garment features at current scale \{$\xi_{i}^{g}(.)$\}. Conditioned on the warped garment feats and person feats \{$\xi_{i}^{p}(.)$\}, the LocalNet predicts local flow \{$f_{i}^{L}$\} and body-parts visbility mask \{$M_{i}^{'}$ \}(which are output to next block). Given garment mask and person representations, GlobalNet predicts global flow \{$f_{i}^{G}$\}. A consistency loss is employed between warped garment masks to harmonize local flows with global alignment.}
\label{fig:warp_block}
\end{figure}

In GC-VTON we use two encoders: (i) \{$\xi^{g}$\} to encode \{$I^{G}$\}, and (ii) \{$\xi^{p}$\} to encode person representations \{$I^{pose}$, $I^{dense}$, and $I^{preserve}$\}. \{$I^{preserve}$\} is the non-garment region extracted from $S$.
The encoded information is then fed to the warping module $\{W_{i}\}_{i=1}^{N}$. Each block of  $\{W_{i}\}$ is composed of a LocalNet \{$W_{i}^{L}$\} and a GlobalNet \{$W_{i}^{G}$\} module (Fig \ref{fig:model}).
LocalNet consumes outputs from \{$\xi^{g}$\} and \{$\xi^{p}$\} to infer fine-grained local flows \{$f^{L}$\} and body parts visibility mask \{$M^{'}$\}.
We add hair and bottom garment mask \{$M^{h+b}$\} from $S$ to \{$M^{'}$\} to form the full body visibility mask \{$M^{vis}$\}.
GlobalNet takes a garment mask \{$M^{G}$\} and output of \{$\xi^{p}$\} to infer global alignment flow \{$f^{G}$\} (Fig \ref{fig:warp_block}). GlobalNet is only needed at train time and is dropped at test time.
Local flow \{$f^{L}$\} is used to warp the original garment \{$I^{G}$\} and the visibility mask \{$M^{vis}$\} is used to mask out occluded regions to generate final warped garment \{$I^{G'}$\}. 
Warped garment \{$I^{G'}$\}, densepose \{$I^{dense}$\}, garment agnostic person image $I^{ag}$ \cite{choi2021viton} and visibility mask \{$M^{vis}$\} are then input to the generator, which generates try-on image \{$I'$\}.

\subsection{Globally Consistent Local Warping}
In a virtual try-on network, garment warping must take care of two tasks: local textures should be preserved and global boundaries of the garment must conform to the target person pose and body. 
We argue that local warping is a garment dependent problem where garment texture must be taken into account to prevent undesired artifacts. While global boundary alignment is inherently garment agnostic and does not care about local textures as the only goal is to match global boundaries.
Therefore, using the same network to solve both the problems together creates a tug-of-war scenario.
Generally, this results in unrealistic warps as the network aligns the global boundaries at the expense of bad local warps or vice versa.
We propose to solve this problem by dedicating separate modules for the local warps and global boundaries alignment. This removes the competition between the two goals and results in smooth and coherent warps. The LocalNet exclusively focuses on generating flows that prevent local texture from artifacts. The local flow is then aligned to the global flow using a consistency loss between the warped garment masks of the local and global flows.
This disentangling takes cares of local adjustments before aligning them to global boundaries, thus resulting in distortion-free warped garments.
Formally,
\begin{equation}
    f_{i}^{L} = W_{i}^{L} (\xi_{i}^{g}(I^{g}) \:\: \odot \:\: \xi_{i}^{p}(I^{pose} \odot I^{dense} \odot I^{preserve}))
\label{eq:local_flow}
\end{equation}
\begin{equation}
    f_{i}^{G} = W_{i}^{G} (\xi_{i}^{gm}(M^{G}) \:\: \odot \:\: \xi_{i}^{p}(I^{pose} \odot I^{dense} \odot I^{preserve}))
\end{equation}

where $i$ represents i-th feature scale, and $f_{i}^{L}$, $f_{i}^{G}$ are the predicted local flow, global flow respectively. $W^{L}$ and $W^{G}$ are the local and global warping modules, while $\xi_{i}^{gm}$ is a garment mask encoder inside $W^{G}$. The $\odot$ represents concat operation. Note that in line with the above discussion, our GlobalNet is a function of garment mask (thus texture agnostic) and the LocalNet on the other hand takes original garment image as input (thus texture and details dependent).
Using the predicted flows, warping is achieved as:
\begin{equation}
    y_{i} = warp(f_{i},x_{i})
\end{equation}

where $warp$ is the grid sampling operator, \{$x_{i}$\} is the input tensor at i-th scale, \{$f_{i}$\} is the flow and \{$y_{i}$\} is the warped output. We warp \{$I_{i}^{G}$\} using \{$f_{i}^{L}$\} to obtain \{$I_{i}^{G'}$\}. Similarly, garment mask \{$M_{i}^{G}$\} is warped using local flow \{$f_{i}^{loc}$\} and global flow \{$f_{i}^{G}$\} to obtain \{$M_{i}^{G'/L}$\} and \{$M_{i}^{G'/L}$\} respectively.
The consistency is then enforced by employing an L1 loss between the warped garment masks:
\begin{equation}
    \Lb_{con,i} = L1(M_{i}^{G'/L} \:,\: M_{i}^{G'/G})
\end{equation}

\subsection{Occlusion Handling}
As discussed, garment occlusions by body parts, e.g hands, cause distortions in the warped garment as flows are inherently not suited to handle the occlusions. To this end, we propose to predict body parts visibility mask which is used to explicitly handle occlusions.
Since body parts visibility mask also depends upon the garment \{$I^{G}$\} and person pose, we task the LocalNet to predict an additional output $M_{i}^{'}$ as:
\begin{equation}
    f_{i}^{L}, M_{i}^{'} = W_{i}^{L} (.)
\end{equation}

where $W_{i}^{L} (.)$ is LocalNet and its inputs from Eq \ref{eq:local_flow}. Hair and bottom garments can also contribute to occlusion, but their visibility does not depend on the upper garment.
Therefore, we directly add hair and bottom garment mask \{$M^{h+b}$\} to \{$M_{i}^{'}$\} to form \{$M_{i}^{vis}$\}. \{$M^{h+b}$\} is obtained from person segmentation map \{$S$\}. The occluded regions are masked out from warped output as:
\begin{equation}
    y_{i} = y_{i} * (1 - M_{i}^{vis})
\end{equation}
where the visibility mask is used to mask out the occluded regions. In the backpropagation step, this setup masks out the gradients for LocalNet, thereby preventing garment tear artifacts.

\begin{figure}
\centering
  \includegraphics[width=0.45\textwidth]{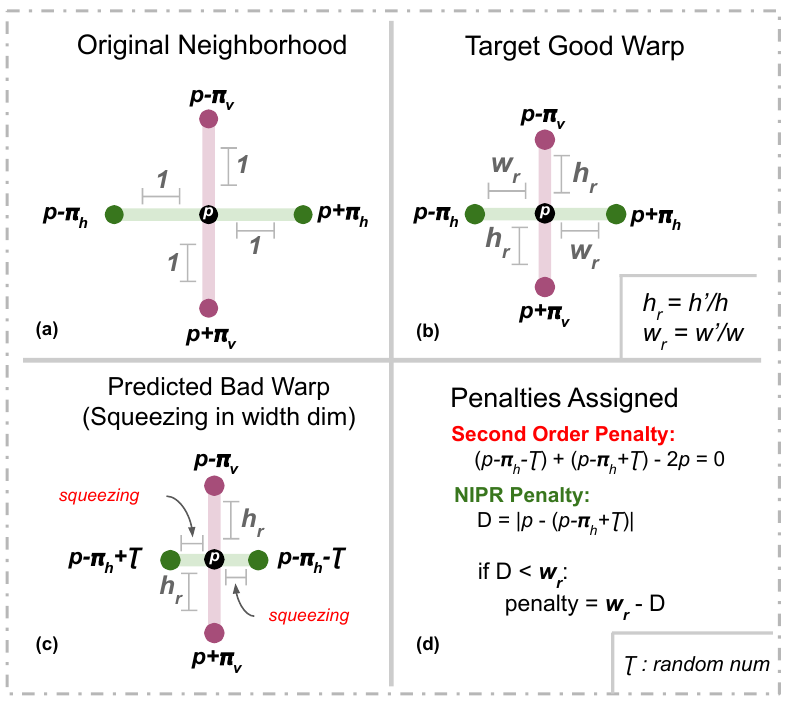}
\caption{NIPR loss illustration. (a) An original local neighborhood. (b) A target true warp that is to be achieved. (c) Predicted bad warp with squeezing artifact in width dimension. (d) Second-order constraint \cite{ge2021parser} is satisfied and fails to correct the bad warp. NIPR appropriately penalizes the flow.}
\label{fig:nipr_motivation}       
\end{figure}

\subsection{Neighborhood Integrity Preserving Regularization (NIPR)}
Unconstrained dense flows try to align the garment to the global pose at the expense of bad local warping. As discussed, in challenging cases such as tucked-in style, the global alignment necessitates predicting high flow values in the bottom region of the garment.
This causes unwanted texture squeezing, which must be guarded against to produce plausible warps.
Existing regularization losses such as Second-Order Smoothness loss \cite{ge2021parser} try to ensure smooth flows by minimzing an objective that encourages equal distances between vertical neighbor pairs of a pixel and similarly for horizontal neighbor pairs (Fig \ref{fig:nipr_motivation}).
But as shown in Fig \ref{fig:nipr_motivation}(d), the configuration is ill-posed as the constraints can be met even when correct warping is not achieved.
This limitation is owed to ignoring global garment changes when equalizing distance in local neighborhoods. This issue is more evident in extreme squeezing and stretching scenarios.
And since the only aim of these losses is to ensure equalized distances between neighbors, they simply cannot identify and correct extreme squeezing and stretching cases.
Formally, the second order smoothness constraint is given as:
\begin{equation}
    \Lb_{so} = \sum_{i=1}^{N} \sum_{p \epsilon P} \sum_{\pi \epsilon [\pi_{v}, \pi_{h}]} \left | f_{i}^{p-\pi} + f_{i}^{p+\pi} - 2f_{i}^{p}\right |
\end{equation}
where $f_{i}^{p}$ is the p-th point on the flow field at i-th scale, and $[\pi_{v},\pi_{h}]$ are the vertical and horizontal neighbors of $p$ (Fig \ref{fig:nipr_motivation}a). The $\left | . \right |$ operator represents absolute function.

In this work, we introduce a novel regularization loss NIPR with the goal of preserving texture integrity in local neighborhoods of the warped garment. NIPR is designed to enforce a simple yet effective principle: distance between pixels in a neighborhood before and after a warp should be consistent and must conform to the global changes.
If distance between two adjacent vertical pixels in the original garment was 'one', then the distance between the same two pixels in the warped garment should roughly be $h_{r}=h{}'/h$. Where $h{}'$ and $h$ are the heights of the warped garment and the original garment respectively.
If the distance is greater than $h_{r}$ then it indicates stretching, if its less, it indicates squeezing (Fig \ref{fig:nipr_motivation}c), and if its equal, it means roughly a perfect warp. Similarly, for two adjacent horizontal pixels, the distance in the warped garment should be $w_{r}=w{}'/w$.

For a particular location, if the distance between two neighbors exceeds $r$ ($r = h_{r}$ for vertical neighbors and $w_{r}$ for horizontal), we simply add this distance to the total loss. This encourages the network to reduce the distance in order to reduce the total loss.
If the distance is less than $r$, we subtract it from $r$ and add the difference to the total loss.
The objective of the loss function now becomes to increase the distance to be equal to $r$. 
When applied to every location of a flow field, NIPR ensures a smooth and coherent warp that is free of the aforementioned artifacts.
Specifically, to formulate NIPR, we add a new component to $\Lb_{so}$:
\begin{equation}
    \begin{aligned}
    \Lb_{nipr} &= \sum_{i=1}^{N} \sum_{p \epsilon P} \Bigg( \Bigg.\sum_{\pi \epsilon [\pi_{v}, \pi_{h}]} \left | f_{i}^{p-\pi} + f_{i}^{p+\pi} - 2f_{i}^{p}\right | \\
    & \quad \quad \quad \quad \quad + \sum_{u \epsilon U}\Lb_{i}^{preserve}(p,u) \Bigg. \Bigg)
    \end{aligned}
\end{equation}
where,
\begin{equation}
    \Lb_{i}^{preserve}(p,u) = \left\{\begin{matrix} D_{i}(p,u) & \: if\,  D_{i}(p,u) > r
     & \\  r - D_{i}(p,u) & \: if\,  D_{i}(p,u) < r
     \\ 0 & otherwise
     & 
    \end{matrix}\right.
\end{equation}

is the integrity preservation component at i-th scale responsible for aligning flows in the local neighborhoods such that they conform to the global changes. $D_{i}(p,u)$ is the absolute distance between a point $p$ on the flow field of i-th scale and its neighbor at a location $u \,\epsilon \,U$. Where $U$ is a set of four neighbors $\{p-\pi_{h}, \, p+\pi_{h}, \, p-\pi_{v}, \, p+\pi_{v}\}$. Please note that the notation for neighbors in $\Lb_{so}$ is different than $\Lb^{preserve}$ because the former works on neighbor pairs and the later deals with a single neighbor at a time. And,
\begin{equation}
    r = \left\{\begin{matrix} h_{r} & \: if \:\: l \,\epsilon\, \pi_{v}
     & \\  w_{r} & \: if \:\: l \,\epsilon\, \pi_{h}
    \end{matrix}\right.
\end{equation}

is the ratio of the warped to the original garment's height or width depending upon the vertical or horizontal neighbor.
Conditioning the penalty on the height and width ratios ensures that the penalty is in line with the global requirements. While conditioning it on the artifact types enables an appropriate response to each artifact, thus alleviating the artifacts effectively.

\subsection{Try-on Generator}
We synthesize the final try-on image $I'$ by utilizing a ResUNet \cite{ronneberger2015u} based encoder-decoder generator $\xi^{gen}$. $\xi^{gen}$ takes the warped garment $I_{N}^{G'}$, dense pose $I^{dense}$, visibility mask $M_{N}^{vis}$, and a clothing agnostic person representation $I^{ag}$ \cite{choi2021viton}. Here $N$ represents the outputs from the last warping block. The visibility mask guides the generator to synthesize skin in the visible regions and at the same time prevents synthesizing garment in the occluded regions.

\subsection{Objective Functions}
To train our warping module, we follow previous works \cite{ge2021parser, he2022style} and utilize $L1$ loss \{$\Lb_{1}^{garment}$\} and perceptual loss \{$\Lb_{p}^{garment}$\} between the warped and the target garments.
We also apply $L1$ loss ($\Lb_{1}^{mask}$) between warped and target garment masks.
In addition, we also use our custom consistency loss \{$\Lb_{con}$\} and NIPR loss \{$\Lb_{nipr}$\}. Since the fine-grained local flows are susceptible to extreme squeezing and stretching, we apply \{$\Lb_{nipr}$\} only to the local flow. To learn body-parts visibility masks, we use binary cross entropy \{$\Lb_{BCE}$\} between the target and predicted masks. So the total objective for the warping module becomes:

\begin{figure*}
\centering
  \includegraphics[width=0.9\textwidth]{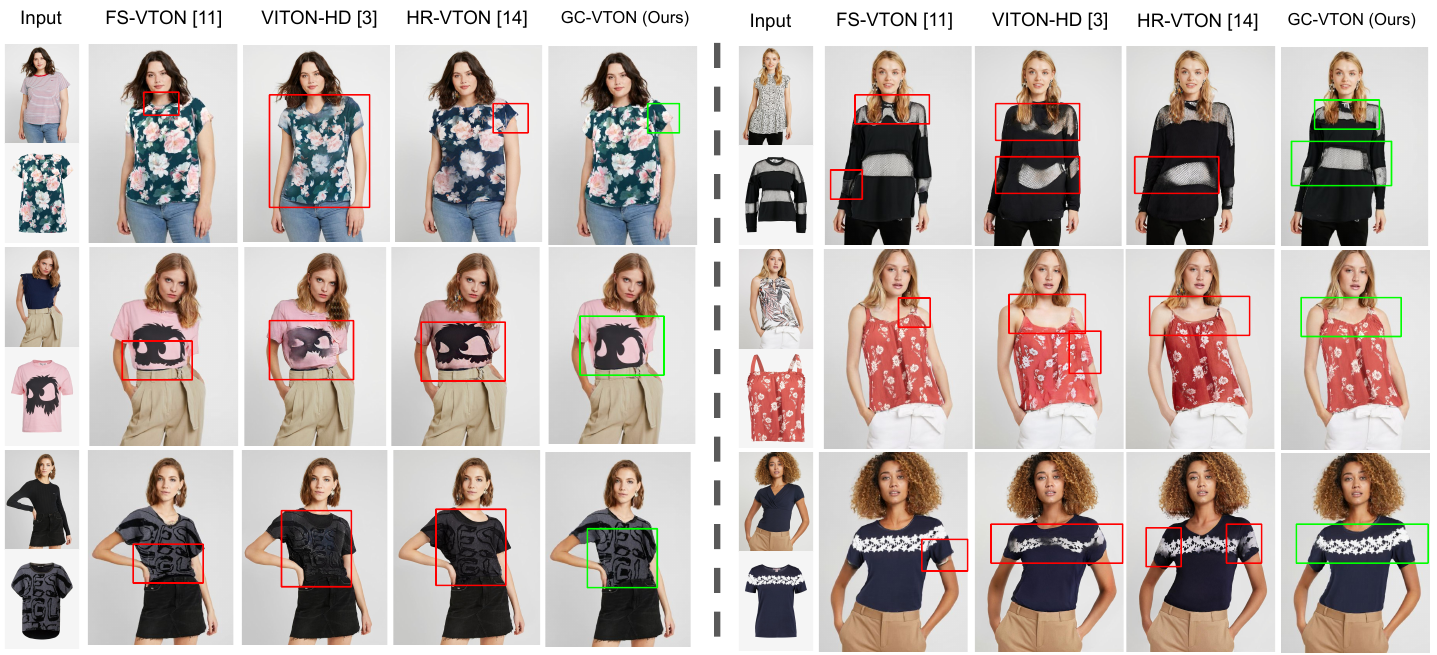}
\caption{Qualitative comparison with other benchmarks. GC-VTON (ours) is able to produce natural looking and artifacts-free try-on images whereas other methods struggle in areas such as texture preservation, texture sharpness, preserving texture linearity and protection against squeezing in tucked-in style. Red boxes show errors in other method and green indicate corrections by our model.}
\label{fig:qualitative_results}       
\end{figure*}

\begin{table}[]
\begin{tabular}{lllll}
\hline
Method         & SSIM$\uparrow$  & FID$\downarrow$    & LPIPS$\downarrow$  & Human$\uparrow$  \\ \hline
VITON-HD       & 0.868 & 9.970  & 0.1141 & 6.72\% \\
HR-VITON       & 0.873 & 9.819  & 0.0954 & 12.78\% \\
FS-VTON        & 0.879 & 7.904 & 0.0951 & 26.2\% \\ \hline
GC-VTON (Ours) & \textbf{0.887} & \textbf{7.888}  & \textbf{0.0831} & \textbf{54.3\%}   \\ \hline
\end{tabular}
\caption{Quantitative comparison to existing methods on VITON-HD \cite{choi2021viton} dataset.}
\label{tab:quantitative}
\end{table}

\begin{equation}
    \begin{aligned}
    \Lb_{warp} &= \{\alpha_{1} \Lb_{1} + \alpha_{2} \Lb_{p}\}^{garment} \\
    &+ \{\alpha_{3} \Lb_{1} \}^{mask} \\
    &+ \alpha_{4} \Lb_{BCE} + \alpha_{5} \Lb_{con} \\
    &+ \alpha_{6} \Lb_{nipr}
    \end{aligned}
\end{equation}

where $\{\alpha_{i}\}_{i=1}^{6}$ are the loss balancing hyper-params for the warping network. 
To train our generator network, we again utilize $L1$ and perception losses for the generated try-on image $I'$.
We also employ an adversarial loss which has proved to be effective in GANs to generate realistic results. The total objective for the Generator network becomes:
\begin{equation}
    \begin{aligned}
    \Lb_{gen} = \{\beta_{1} \Lb_{1} + \beta_{2} \Lb_{p}\}^{try-on} + \beta_{3} \Lb_{adv}
    \end{aligned}
\end{equation}

where $\{\beta_{i}\}_{i=1}^{3}$ are the loss balancing hyper-params for the generator network.
\section{Experiments}
\subsection{Dataset}
For our work, we use the widely used virtual try-on dataset VITON-HD \cite{choi2021viton}. It contains 13,769 frontal-view female and upper garment pairs for training. The testing set contains 2,032 person and garment pairs. The garment in a training pair is the same that the person is wearing. For a testing pair, the garment is different than what the person is wearing. The original dataset has an image resolution of 1024x768 and we resize these images to 256x192 for our warping module and 512x384 for the generator module.

\begin{figure}[h]
\centering
  \includegraphics[width=0.45\textwidth]{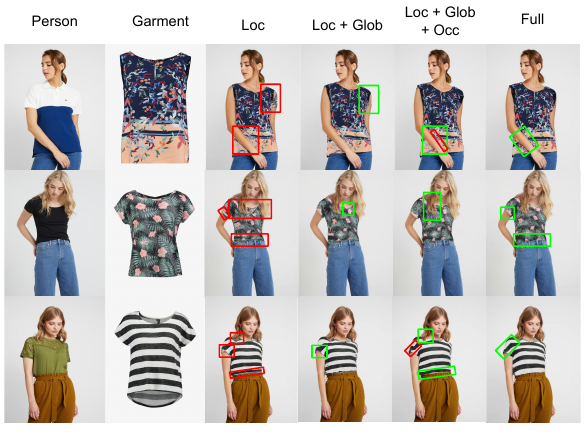}
\caption{Qualitative comparison of different components of our pipeline. Green boxes show progressive improvement as each component is added.}
\label{fig:ablation_qualitative}       
\end{figure}

\subsection{Implementation details}
Our model is trained with a single RTX 3090 12 GB GPU. Our warping module is a set of $N$ Conv-LeakyReLU blocks ($N=5$ in this work). We train the warping module for 100 epochs with a batch size of 14 at 256x194 resolution. The output flow from warping module is upscaled by a factor of two to warp the garment at a resolution of 512x384, which is the input and output resolution for the generator network. The generator module is trained for 100 epochs with a batch size of four and is based on  the implementation of \cite{he2022style}. 
We use Adam optimizer with an initial learning of $5e-4$ which linearly decays after 50 epochs. The loss mixing hyper-params for warping modules training are empirically found out and set to $\{\alpha_{i}\}_{i=1}^{6} = \{1, 0.2, 2, 2, 1, 1\}$. The generator loss mixing hyper-params are set to $\{\beta_{i}\}_{i=1}^{3} = \{1, 5, 1\}$.

\subsection{Evaluation metrics and Baseline Methods}
As per standard practice, we use Structurual Similarity Index (SSIM) \cite{wang2004image}, Perceptual Distance (LPIPS) \cite{zhang2018unreasonable} and Fr\'echet Inception Distance (FID) \cite{parmar2022aliased} to evaluate the quality of the generated images. SSIM and LPIPS are employed in a paired fashion whereas the FID is used in unpaired settings. Additionally, we also conduct a human study where we show 20 random images from each baseline method to each participant in a 10 person pool. The participants rate the images produced by each method in order of realism and we report the percentage of the preference for each method.

We compare our results to the current SOTA methods including FS-VTON \cite{he2022style}, HR-VTON \cite{lee2022high}, and VITON-HD \cite{choi2021viton}. For all the methods we directly use their official codes to generate try-on images. For FS-VTON \cite{he2022style}, we train their network from scratch as they don't have results available on VITON-HD dataset. We re-calculate the scores for each method using the obtained results.

\begin{table}[]
\begin{tabular}{llll}
\hline
Method           & SSIM$\uparrow$  & FID$\downarrow$    & LPIPS$\downarrow$  \\ \hline
Loc              & 0.8729 & 8.236  & 0.1032 \\
Loc + Glob       & 0.8754 & 8.168  & 0.0912 \\
Loc + Glob + Occ & 0.882 & 8.003 & 0.0899 \\ \hline
GC-VTON (Full)   & \textbf{0.887} & \textbf{7.888} & \textbf{0.0831} \\ \hline
\end{tabular}
\caption{Quantitative comparison of all the components of our method.}
\label{tab:ablation_quantitative}
\end{table}

\subsection{Results}
The quantitative results of our model are given in Table \ref{tab:quantitative}. Our GC-VTON clearly outperforms all the previous works on all the evaluation metrics. This confirms that our method consistently produces realistic try-on images compared to the baselines. Additionally, human evaluators also prefer the output of our method 54.3\% of the times when presented side-by-side with other methods. A detailed qualitative comparison of GC-VTON (ours) against the existing works is presented in Figure \ref{fig:qualitative_results}. Existing models run into various artifacts in the generated images like missing key details from original garment (Lr1c4, Rr2c2, Rr2c3, Rr2c4) $\{Lr2c2 = Left\: Row2\: Col2\}$, unrealistic hair fusion (Lr2c3, Rr1c3), failing to preserve garment texture (Lr2c3, Lr3c3, Lr3c4, Rr3c3), unable to preserve texture around occlusions from body parts (Rr1c2), texture squeezing and stretching (Lr2c3, Lr3c2, Lr3c3), misalignment in the sleeves texture (Lr1c4, Rr3c4) and failing to maintain linearity of the texture (Rr1c2, Rr2c3, Rr1c4). In contrast our model does not suffer from these issues. We have to thank Local-Global disentanglement for aligning global boundaries and preserving local texture (Lr1c5, Lr3c5). Occlusion handling through visibility mask efficiently handles artifacts around occlusions (Rr1c5). Employing NIPR brings texture linearity preservation (Rr1c5) and preventing garments from extreme squeezing in tucked-in style (Lr2c5, Lr3c5). Additional results can be seen in supplementary materials.

\subsection{Ablation Studies}
In this section we present a qualitative (Figure \ref{fig:ablation_qualitative}) and quantitative analysis (Table \ref{tab:ablation_quantitative}) of our method and show how each component of our method plays an integral part in obtaining realistic try-on results. Multiple experiments are conducted to progressively add and evaluate the components. Here \{Loc\} refers to using LocalNet only, \{Loc + Glob\} refers to LocalNet + GlobalNet, \{Loc + Glob + Occ\} represents LocalNet + GlobalNet + Occlusion Handling and finally \{Full\} refers to full GC-VTON network with NIPR loss. In Figure \ref{fig:ablation_qualitative}, red boxes show errors and green boxes show how addition of each component helped mitigate the errors one by one.

\textbf{Use of global consistency: }
demonstrates improvements in the SSIM $(0.8729 \rightarrow 0.8754)$, a drop in FID ($8.236 \rightarrow 8.168$) and LPIPS ($0.1032 \rightarrow 0.0912$) as shown in Tab \ref{tab:ablation_quantitative}. Qualitative evidence of GlobalNet's positive role is shown in Fig \ref{fig:ablation_qualitative} col-4, where in the row-1 the global alignment of left shoulder is corrected. Row-2 indicates the neck region misalignment in \{Loc\} which is effectively resolved by the global consistency. In row-3, adding GlobalNet helped fix the right sleeve misalignment. The effective resolution of global boundary alignment at multiple garment locations is indicative of the strong abilities of the global consistency loss.

\textbf{Occlusion Handling:}
Consistent improvements in SSIM, LPIPS and FID scores are observed when occlusion is explicitly handled. As indicated previously, handling all possible occlusion sources is vital for a distortion-free realistic warp. Results in Fig \ref{fig:ablation_qualitative} col-5 are a visual confirmation that our occlusion handling mitigates occlusions induced by various sources. Specifically in row-1, distortions caused by hands are mitigated. In row-2 occlusion by hair is handled while in row-3 the squeezing due to tucked-in style (occlusion by another garment) is handled. 

\begin{figure}
\centering
  \includegraphics[width=0.38\textwidth]{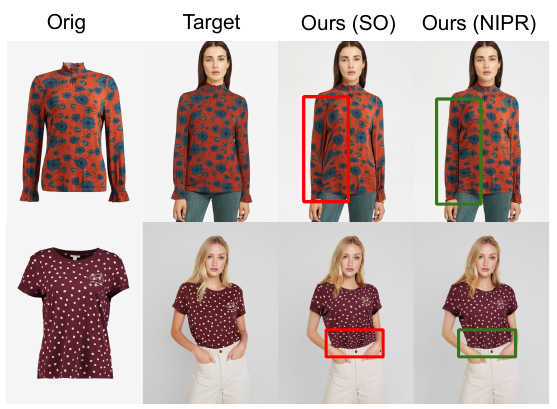}
\caption{NIPR vs Second-Order constraint qualitative comparison. NIPR is a better guard against artifacts than the existing regularization losses. Please see supplementary for more results.}
\label{fig:nipr_qualitative}       
\end{figure}

\textbf{NIPR: }
is an integral component of our pipeline and quantitative results in Table \ref{tab:ablation_quantitative} confirm the perceptual improvements it brings. Visual analysis in Fig \ref{fig:ablation_qualitative} column 6 confirms the intuition behind the loss. In row-1 and row2, the left-over squeezing  around occluded regions is alleviated. In row-3, the linearity of the the lines on the right sleeve is preserved, confirming the co-linearity preservation of local neighborhoods. 

To dig a bit deeper, we train a model \{Ours(SO)\} with all our components, except the NIPR is replaced with the second-order (SO) smoothness loss. In order to have access to a target warp, we compare the performance of this model in a paired setting to our model with NIPR. Fig \ref{fig:nipr_qualitative} row-1 suggests that the model trained with Second Order constraint suffers from extreme texture stretching compared to the target warp. In the second row, the model with SO constraint suffers from extreme squeezing due to tucked-in style while NIPR effectively guards against the artifacts in both the cases. This is due to the fact that NIPR can identify the artifacts and appropriately assign penalties to the predicted flows. While the other losses only focus on local windows, NIPR is aware of global changes and tames the local neighborhoods to be in line with the global changes.

\section{Conclusion}
In this work, we present a novel method to generate artifacts-free and natural try-on images. For warping the garment via a flow field, we disentangle the global boundary alignment task from local texture preservation, thus allowing the network to equally focus on both. The outputs of the two tasks are matched via a consistency loss, thus harmonizing the local and global flows. To prevent artifacts around occluded regions, we predict a body-parts visibility mask, which masks out the occluded regions in the warped garment to prevent flows from creating distortions. Lastly, we propose a novel flow regularization loss NIPR that penalizes bad flows in local neighborhoods. NIPR applies a penalty that is carefully designed to appropriately tackle artifacts and ensure that local neighborhoods conform to global changes in the garment. Evaluation on VITON-HD dataset shows strong performance compared to the baselines both qualitatively and quantitatively.


{\small
\bibliographystyle{ieee_fullname}
\bibliography{egbib}
}

\end{document}